\documentclass[sigplan,nonacm]{acmart}

\usepackage{algorithm}
\usepackage{array}
\usepackage{textcomp}
\usepackage{stfloats}
\usepackage{url}
\usepackage{verbatim}
\usepackage{graphicx}
\usepackage{balance}
\usepackage{tikz}
\usepackage{algpseudocode} 
\usepackage[acronym]{glossaries}
\makeglossaries

\newcommand{\ours}{SpecPipe}
\newcommand{\oursdb}{SpecPipe-DB}
\newacronym{llm}{LLM}{large language model}
\newacronym{tbt}{TBT}{time between tokens}
\newacronym{kvcache}{KV Cache}{Key-Value Cache}
\newacronym{pp}{PP}{pipeline parallelism}
\newacronym{tp}{TP}{tensor parallelism}

\usepackage{tabularx,booktabs,array}
\usepackage{pifont} 
\usepackage{adjustbox}

\newcommand{\cmark}{\ding{51}}
\newcommand{\xmark}{\ding{55}}

\begin{document}

\title{{\ours{}}: Accelerating Pipeline Parallelism-based LLM Inference with Speculative Decoding}

\author{Haofei Yin}
\affiliation{%
  \institution{Shandong University}
  \city{}
  \country{}
}

\author{Mengbai Xiao}
\affiliation{%
  \institution{Shandong University}
  \city{}
  \country{}
}
\email{xiaomb@sdu.edu.cn} 

\author{Tinghong Li}
\affiliation{%
  \institution{Shandong University}
  \city{}
  \country{}
}

\author{Xiao Zhang}
\affiliation{%
  \institution{Shandong University}
  \city{}
  \country{}
}

\author{Dongxiao Yu}
\affiliation{%
  \institution{Shandong University}
  \city{}
  \country{}
}

\author{Guanghui Zhang}
\affiliation{%
  \institution{Shandong University}
  \city{}
  \country{}
}

\begin{abstract}
  The demand for large language model inference is rapidly increasing. Pipeline parallelism offers a cost-effective deployment strategy for distributed inference but suffers from high service latency. While incorporating speculative decoding to pipeline parallelism improves performance, it still faces challenges of low hardware utilization and narrow speculative window. Inspired by branch prediction in instruction pipelining, we introduce \ours{}, which fills the pipeline with speculative tokens of a request step-by-step. By maximizing the hardware utilization, \ours{} decodes one token per pipeline step ideally. Specifically, \ours{} comprises a dynamic speculative token tree and a pipelined inference framework. The tree dynamically accepts tokens from a speculative token source and outputs the tokens to the inference pipeline. Since the speculative window relaxed in our framework, a high-accuracy draft model is integrated without fine-tuning. The pipeline inference framework follows node-wise computation, pruning propagation, and inter-node communication stages. We implement \ours{} and a variant \oursdb{} with dynamic batching for single- and multi-request inference, respectively. On an 8-stage pipeline, \ours{} improves time between tokens on diverse single-request workloads by $4.19\times$–$5.53\times$ over standard pipeline parallelism and by $2.08\times$–$2.38\times$ over prior tree-based speculative decoding methods. For multi-request workloads, \oursdb{} achieves $1.64\times$–$2.08\times$ higher throughput and $1.61\times$–$2.06\times$ lower time between tokens than vLLM. 
\end{abstract}

\maketitle


\section{Introduction}
\label{sec:introduction}

The rapid development of \glspl{llm} has shown remarkable potential across various domains~\cite{Hou2023Large, Omiye2023Large, motlagh2024large, topsakal2023creating}, and the demand for efficient inference continues to grow. As \glspl{llm} have scaled to trillions of parameters~\cite{brown2020language, zhang2022opt, dubey2024llama, wagh2020falcon}, even the highest-end GPU equipped with HBM3/4~\cite{Kim202530.3, Kim2024Present} can hardly contain an entire model in one device. Thus, distributed inference has become the de facto approach for deploying \glspl{llm}.

\begin{figure}[t]
  \centering
  \includegraphics[width=\columnwidth]{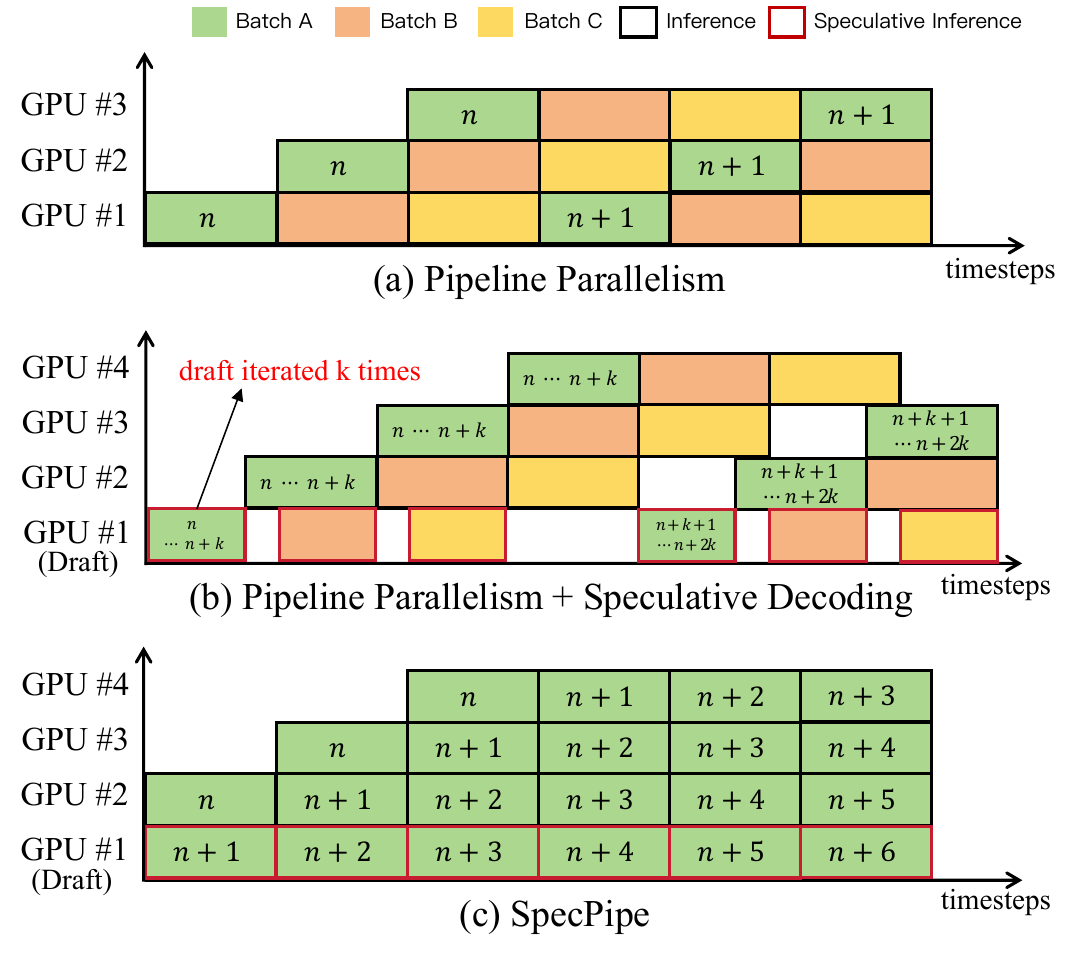}
  \caption{The pipeline timing diagram of \gls{llm} inference with (a) \gls{pp}, (b) \gls{pp}${+}$speculative decoding, and (c) \ours{} on 3 GPUs. \gls{pp} and \gls{pp}${+}$speculative decoding infers 3 batches for optimizing hardware utilization. \gls{pp}${+}$speculative decoding and \ours{} deploy the draft model on an additional GPU. All methods start from inferring the $n$-th token. In \gls{pp}${+}$speculative decoding, $k$ tokens are predicted every inference pass.}
  \label{fig:intro}
\end{figure}

However, the communication and synchronization overhead of distributed inference inherently increases service latency~\cite{xu2025characterizing, butler2024pipeinfer}. This hinders the deployment of latency-sensitive \gls{llm}-based services, such as real-time dialogue systems and completion tools.
To effectively alleviate the communication overhead, {\it \gls{tp}}~\cite{shoeybi2019megatron} has to employ high-speed yet expensive interconnects like NVLink or InfiniBand~\cite{li2019evaluating}. This suggests that \gls{tp} becomes impractical in scenarios where such technologies are infeasible or not affordable, such as when GPUs are distributed across different racks, data centers, or edge computing environments, where {\it \gls{pp}} is favored~\cite{Agrawal2024TamingTT, hu2021pipeline, ma2024hpipe, butler2024pipeinfer}. \gls{pp} offers a more cost-effective deployment strategy, as communication occurs only once across several layers. 
Nevertheless, the latency of \gls{pp} inferring one request, compared to \gls{tp}, remains high due to its low hardware utilization: the pipelined GPUs are sequentially activated so that only one device is working at any given moment for a single request. Multi-batch inference improves the hardware utilization of \gls{pp} by feeding different requests into the pipeline each step, as illustrated in Figure~\ref{fig:intro}(a). While this approach effectively optimizes inference throughput of \gls{pp}, it hardly reduces or even increases inference latency~\cite{Agrawal2024TamingTT, Guo2025gLLMGB}.

Speculative decoding~\cite{xia2024unlocking, zhou2023distillspec, liu2023online, li2024eagle, li2024eagle2, cai2024medusa, zhang2023draft, spector2023accelerating, miao2024specinfer} is a promising technique for reducing inference latency. It employs a lightweight draft model to quickly speculate a series of future tokens, then feeds the speculative tokens to the large inference model at once. If all predictions are correct, multiple tokens are decoded in one step, significantly accelerating inference. However, directly incorporating speculative decoding into \gls{pp}, illustrated in Figure~\ref{fig:intro}(b), presents challenges: 1) For single-request inference, the approach still activates one GPU per pipeline stage, fundamentally limiting its acceleration potential. 2) For multi-request inference, the speculative window is constrained to a single step, implying that only small and potentially inaccurate draft models are deployable. Whenever a misprediction occurs, the inference latency reverts to that of standard pipeline parallelism in the worst case, and computational resources are wasted on processing incorrect tokens. 3) The draft model speculates future tokens only after the previous round of tokens have been inferred, lowering the hardware utilization.

We propose a novel method to integrate speculative decoding into \gls{pp}, effectively reducing its inference latency. Inspired by branch prediction in instruction pipelining, we feed speculative tokens of a batch into the pipeline step-by-step rather than all at once, as illustrated in Figure~\ref{fig:intro}(c). This effectively addresses the challenges of naively integrating speculative decoding into \gls{pp}. First, by feeding speculative tokens at each step, all GPUs remain active even for single-request inference. Second, the speculative window is significantly relaxed, requiring token prediction at only one position per step; this allows a larger and more accurate draft model to be deployed, substantially improving prediction accuracy as well as the system performance. Third, the token speculation via draft model can be covered by other pipeline steps. In the best case, this method decodes one token per pipeline step.

In this paper, we present \ours{}, a speculative decoding framework that accelerates pipeline-based LLM inference. \ours{} is composed of two components: {\it a dynamic speculative token tree} and {\it a pipelined inference framework}. During the inference, a speculative token tree is dynamically maintained, whose root is the last inferred token and each tree level representing a future token position. As the tree branches, multiple candidate tokens are predicted for each position. At each pipeline step, a tree-level of speculative tokens is fed to the pipeline from one end, and the other end successfully infers a token. The tree is updated via two operations: {\it layer appending} adds speculative tokens for the next position, and {\it to-subtree pruning} reroots the tree once a speculative token is verified. The speculative tree accepts speculative tokens from a token source, which guarantees that at least a tree level is generated per pipeline step. Since the speculative window is relaxed, a large pre-trained draft model becomes deployable, pushing the prediction accuracy of next speculative tokens beyond 95\% without any fine-tuning. Our pipelined inference framework has three stages in a pipeline step: 1) it first carries out node-wise computation to calculate token embeddings and to infer the next token across devices, 2) then the inferred token is used to prune the speculative tree, the respective embeddings, and \gls{kvcache} on devices, i.e., the pruning propagation stage. 3) In the inter-node communication stage, the remaining speculative embeddings are delivered to the next device in the pipeline. To optimize performance, the speculative tree is configured to be fixed-width for balancing workloads on all devices. The tree width selection is subtle: a small width shortens step time but also increases the chance of pipeline flushes. Our analysis suggests that a proper tree width should be carefully selected to minimize \gls{tbt}.

We implement \ours{} for single-request inference and a variant \oursdb{} optimized towards multi-request workloads. We deploy \ours{} on an 8-stage pipeline running the inference model of LLaMA3.1-70B, with a dedicated GPU hosting LLaMA3.2-1B as the draft model. Experimental profiling shows that a tree width of 64 minimizes \gls{tbt} on our testbed. In single-request inference, \ours{} achieves the lowest \gls{tbt}, outperforms vanilla PP, PP-vLLM~\cite{kwon2023efficient}, SpecInfer~\cite{miao2024specinfer}, and EAGLE3~\cite{li2025eagle} by \(4.19\times\)--\(5.53\times\), \(2.50\times\)--\(3.32\times\), \(2.09\times\)--\(2.38\times\), and \(3.58\times\)--\(4.98\times\), respectively. A similar performance gain is observed on a two-server configuration, where \ours{} outperforms the tensor parallelism implemented in vLLM by \(2.55\times\)--\(3.44\times\).  In multi-request experiments, the \oursdb{} delivers \(1.64\times\)--\(2.08\times\) higher throughput and \(1.61\times\)--\(2.06\times\) lower \gls{tbt} than vLLM.

Our contributions are as follows:
\begin{itemize}
\item We propose \ours{}, a pipelined \gls{llm} inference system powered by speculative decoding. By feeding speculative tokens step-by-step, it maximizes hardware utilization even in single-request inference and sharply reduces \gls{tbt} nearly one token per step.
\item We design a GPU-oriented speculative tree that supports dynamic extension and pruning. The tree coordinates the speculative token source and the pipelined inference framework, and it timely removes mispredicted tokens to avoid redundant work.
\item We realize a pipelined inference framework that utilizes a highly accurate draft model as the token source, whose speculative latency can be covered by other pipeline steps. The framework owns a complete stepping mechanism and optimizations for pipelining performance.
\item We implement \ours{} and its multi-request variant \oursdb{}. \ours{} outperforms the state-of-the-art speculative decoding in \gls{pp} by \(2.09\times\)--\(2.38\times\). \oursdb{} reaches \(1.64\times\)--\(2.08\times\) higher throughput and \(1.61\times\)--\(2.06\times\) lower \gls{tbt} than vLLM.
\end{itemize}

The remainder of this paper is organized as follows: Section~\ref{sec:background} provides the background. Section~\ref{sec:design} presents the system design. Implementation details and evaluation results are discussed in Section~\ref{sec:evaluation}. Section~\ref{sec:RelatedWork} reviews related work. Finally, Section~\ref{sec:ConclusionAndFutureWork} concludes the paper.

\section{Background}
\label{sec:background}

\subsection{\gls{llm} Inference}
\glspl{llm} are transformer-based networks with tens to hundreds of billions of parameters. Given a prompt (a token sequence), the model autoregressively generates the next token until a stop condition such as {\tt [EOS]}. Inference has two stages: a \emph{prefill} stage that processes the prompt and produces the first new token, and a \emph{decoding} stage that generates the remaining tokens one by one.

\glspl{llm} used in this work are decoder-only stacks of repeated layers, each consisting of a self-attention module followed by a feed-forward network (FFN)~\cite{radford2018improving, brown2020language}. In self-attention~\cite{vaswani2017attention}, given $\mathbf{Q},\mathbf{K},\mathbf{V}\in\mathbb{R}^{n\times d}$, the attention matrix is computed as \(\mathbf{S}=\text{softmax}\left(\mathbf{Q}\mathbf{K}^\top/\sqrt{d}\right)\) and the attention output is $\mathbf{O}=\mathbf{S}\mathbf{V}$. For text generation, a causal mask restricts attention to preceding tokens. When a new token is processed, only one additional row is computed at each layer, while the stored $\mathbf{K}$ and $\mathbf{V}$ from previous tokens are reused through a \gls{kvcache}. Figure~\ref{fig:attention_masks}(a) illustrates this computation.

After the attention and FFN computations, the model outputs a categorical probability distribution over all possible tokens at the next position. Under a greedy decoding policy, the token with the highest probability is selected as the output. For stochastic decoding, the next token is drawn from a temperature-scaled probability distribution that can be restricted using top-$k$ or top-$p$ sampling. In top-$k$ sampling, only the $k$ most probable tokens are retained as candidates. In top-$p$ (nucleus) sampling, the candidate set is the smallest group of tokens whose cumulative probability meets or exceeds a predefined threshold $p$. Probabilities are then renormalized within this set before sampling. A lower temperature makes the probability distribution more peaked, favoring high-probability tokens and producing more deterministic outputs, while a higher temperature flattens the distribution and increases output diversity.

\subsection{Speculative Decoding}
\begin{figure}[t]
  \centering
  \includegraphics[width=\columnwidth]{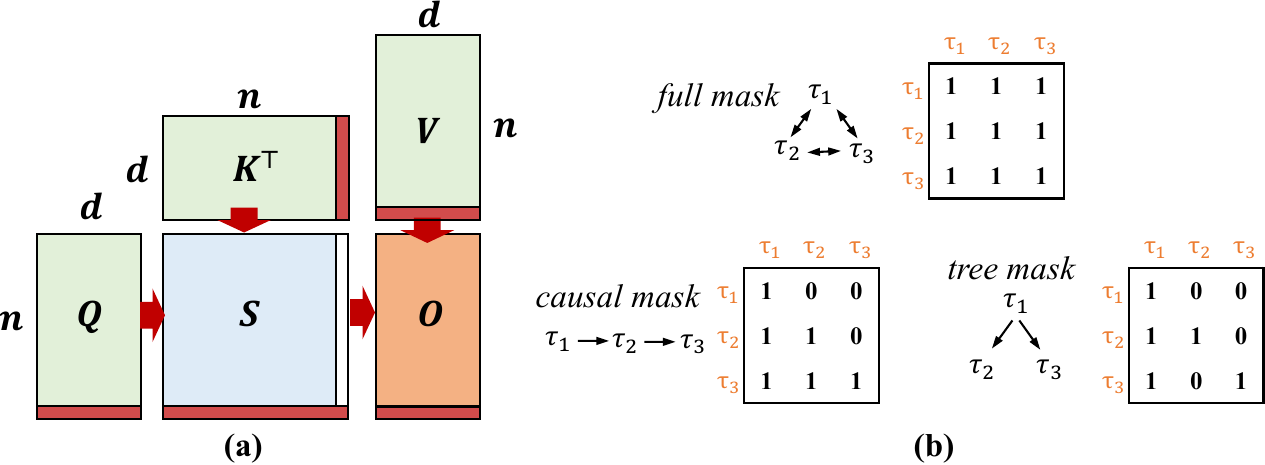}
  \caption{(a) The computation of self-attention. The red area means that when a causal mask is applied, adding a new token only requires computation of a new row. (b) Different attention masks including full mask, causal mask, and tree mask. $\tau_i$ are tokens.}
  \label{fig:attention_masks}
  \small
\end{figure}

Speculative decoding~\cite{xia2024unlocking, zhou2023distillspec, liu2023online, li2024eagle, li2024eagle2, cai2024medusa, zhang2023draft, spector2023accelerating, miao2024specinfer} is a technique for accelerating LLM inference. Under the autoregressive decoding pattern, tokens are generated sequentially. Most time is spent on loading model parameters from GPU memory to registers, despite the small per-token computation. To more effectively utilize hardware resources, one could ``speculate'' tokens at the next positions before the inference, and verify them in one pass of inference. For example, as we already have $n$ tokens, we speculate a few tokens at the positions from $n{+}1$ to $n{+}k$ prior to inference. After one pass of inference, the inference model $(n{+}1)$-th token is inferred. If this token matches one of the speculative tokens at the $(n{+}1)$-th position, we can immediately get the true token at the $(n{+}2)$-th position since the speculative tokens are also involved in the inference. With a chained verification process, we may accept up to $k$ tokens in an inference pass. The speculative tokens could be organized into a chain that allows one token at a position, or into a tree where multiple tokens are guessed for a position. When organizing speculative tokens in a tree, the causal mask has to be modified to a {\it tree mask} to enforce valid attention dependencies under tree-structured speculation. Figure~\ref{fig:attention_masks}(b) illustrates an example of such a tree mask.

In speculative decoding, the time of generating speculative tokens and the prediction accuracy are the keys. If the speculative token generation takes too long, it offsets the gain of inferring multiple tokens simultaneously. If the prediction accuracy is low, we could spend more computation power but still infer a token in a pass. A common solution is that a small model is trained to mimic the original model for generating speculative tokens. We call the small model a {\it draft model}, and the original model the {\it inference model}. There are also studies~\cite{fu2024break,he2023rest,hu2024sam} generating speculative tokens from the prefilled tokens or from a large database.

\section{\ours \ Design}
\label{sec:design}

\begin{figure*}[t]
  \centering
  \includegraphics[width=0.95\textwidth]{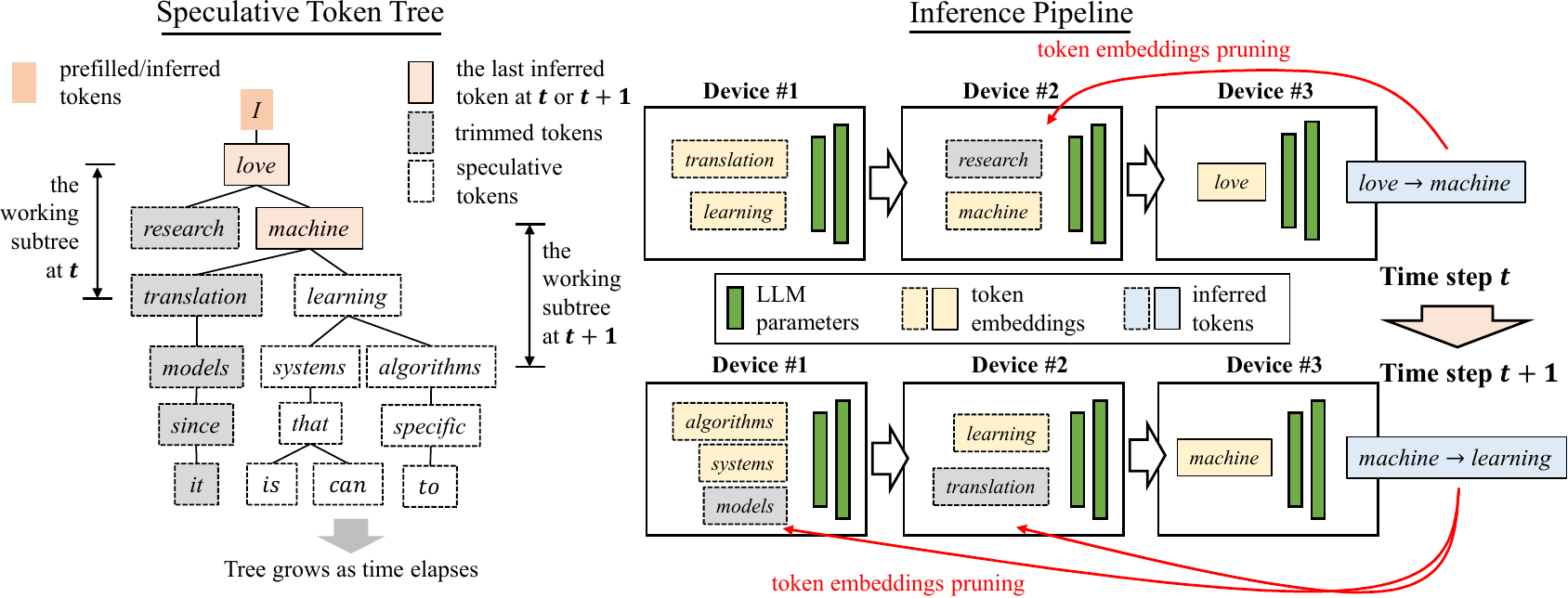}
  \caption{Overview of \ours{}. The framework employs three pipelined devices, each hosting several LLM layers. At the current step, the input has advanced to {\it ``I love''}. The left panel shows the speculative token tree extending from {\it ``I love''}, while the right panel depicts device states at time steps $t$ and $t{+}1$, where each device processes one tree level per step. At $t$, \ours{} selects {\it ``machine''} after {\it ``love''}, rerooting the tree to {\it ``machine''} and pruning the subtree rooted at {\it ``research''} along with its embeddings (pruned before transmission). For clarity, the \gls{kvcache} is omitted.}
  \label{fig:overview}
  \small
\end{figure*}

\begin{figure}[t]
  \centering
  \includegraphics[width=0.9\columnwidth]{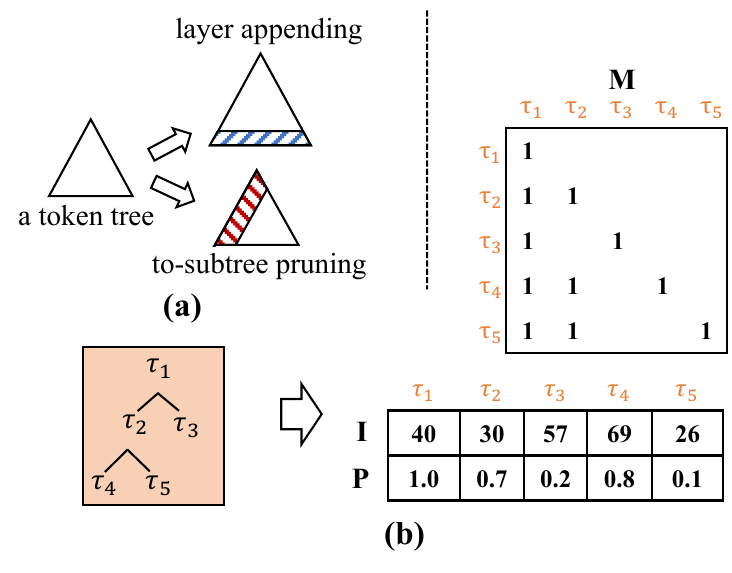}
  \caption{(a) Our dynamic tree supports two operations: layer appending and to-subtree pruning. (b) An example speculative tree of five tokens.}
  \label{fig:tree_presentation}
  \small
\end{figure}

\subsection{System Overview}
\ours{} is a pipelined inference framework for large language models (LLMs). In our design, speculative decoding is integrated to improve hardware utilization and lower the \gls{tbt} in the {\it decoding} stage, thus reducing the overall inference latency of a given query. Figure~\ref{fig:overview} shows how \ours{} processes an inference query during the decoding stage.

As shown in Figure~\ref{fig:overview}, \ours{} feeds the pipelined inference framework with tokens from a speculative token tree. Unlike the conventional speculative decoding solutions which generate a static tree and input it at once~\cite{miao2024specinfer}, our tree is gradually generated and fed to the pipeline per step. Once the pipeline is filled, \ours{} is expected to output a token per pipeline step. Specifically, at an arbitrary timestep $t$, we always have a speculative tree whose root token embedding resides in the last device, and the next token is inferred. If the inferred token is one of the speculative ones, it helps prune irrelevant embeddings across devices. 
At the end of a pipeline step, the remaining tokens are delivered to the next device. There is a risk that the inferred token does not exist in the speculative tree, which causes the speculative tree to be discarded and all embeddings in the pipeline to be removed. As a result, some stages may become idle or the entire pipeline may be flushed, requiring inference to restart from the current position and degrading latency to that of standard pipeline parallelism.

Next, we will first introduce how we maintain the {\it dynamic speculative tree}, and then based on this, how the {\it pipelined inference framework} works and is optimized.


\begin{figure*}
  \centering
  \includegraphics[width=0.95\textwidth]{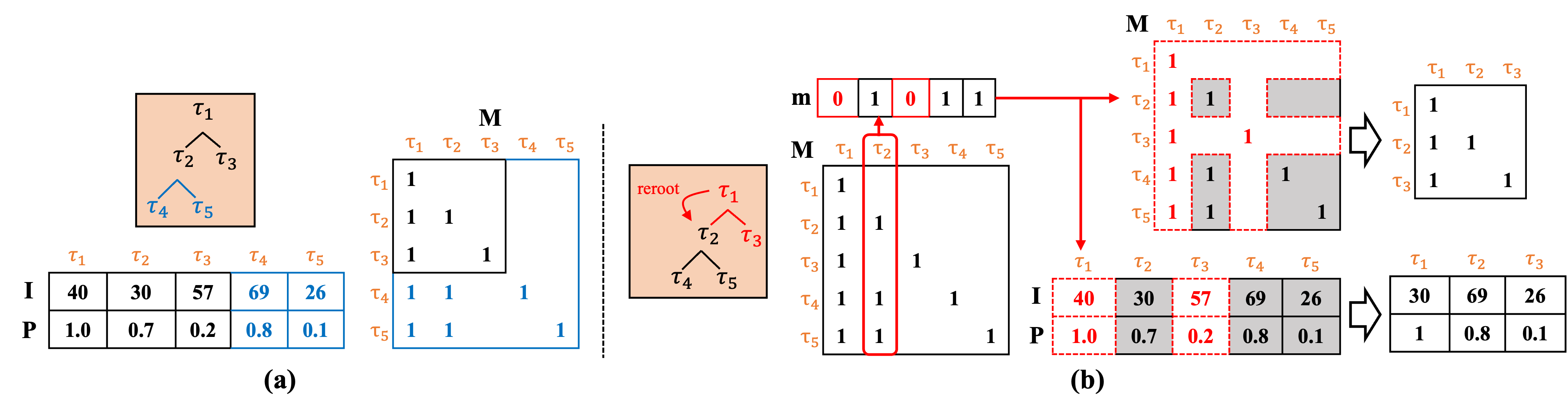}
  \caption{Two updating examples of our dynamic speculative token tree: (a) {\it layer appending} and (b) {\it to-subtree pruning}.}
  \label{fig:tree_updating}
  \small
\end{figure*}

\subsection{Dynamic Speculative Tree}
\label{sec:dst}

\subsubsection{Tree presentation and operations}
A speculative token tree for an inferring query is composed of the last verified token as the root and the predicted tokens as intermediate nodes and leaves. The tree height $h$ indicates that $h{-}1$ future positions are predicted with the tokens at the respective tree levels. As the auto-regressive inference is proceeding, we must periodically 1) add nodes to the tree because new speculative tokens are generated, and 2) reroot the tree to one of its subtrees since some of the speculative tokens are verified. Thus, our dynamic tree supports two essential updates: {\it layer appending} and {\it to-subtree pruning}, which are depicted in Figure~\ref{fig:tree_presentation}(a). To realize the update operations efficiently, a token index $\mathbf{I}$, a probability array $\mathbf{P}$, and a mask matrix $\mathbf{M}$ are created to represent the tree.

Specifically, if we have a tree composed of $n$ tokens, the tokens are indexed from root to leaves by a Breadth-First Search (BFS) order, so that $\tau_1$ is always the root. In the token index array, $\mathbf{I}_i$ denotes the token index of $\tau_i$. $\mathbf{P}_i$ contains the probability of $\tau_i$ generated from its parent node in the tree. For $i > j$, $\mathbf{M}_{i,j} = 1$ indicates that $\tau_i$ is a successor of $\tau_j$ in one of the predicted sequences. When $i = j$ or $i < j$, $\mathbf{M}_{i,j}$ is always 1 or 0. 
Note that $\mathbf{M}$ is also used in self-attention computation. Following~\cite{miao2024specinfer}, we use a tree mask to enable batched speculative decoding. However, unlike their static formulation, we maintain a dynamic tree mask matrix to support pruning and updates during inference. Figure~\ref{fig:tree_presentation}(b) shows an example of our tree representation.

\noindent{\bf Layer appending}. Whenever \ours{} generates a number of speculative tokens for a new position, we append these tokens to the tree bottom as a new layer. Since tokens are indexed level by level, carrying out layer appending is straightforward: after sorting the new tokens according to the order of their parent nodes, they are appended to $\mathbf{I}$ and $\mathbf{P}$. $\mathbf{M}$ is also expanded, where the old mask matrix is the top-left area. The bottom-left area is set based on the predecessors of new tokens, the bottom-right area is an identity matrix, and the top-right area is an empty matrix. Figure~\ref{fig:tree_updating}(a) shows an example of adding two new tokens to a speculative tree with three tokens.

\noindent{\bf To-subtree pruning}. Moving the root to a subtree means that a substantial portion of the original tree is pruned. Relying on the mask matrix $\mathbf{M}$, this could be implemented efficiently. Once a new root $\tau_i$ is determined, we extract the $i$-th column of $\mathbf{M}$ to a temporary mask array $\mathbf{m}$. $\mathbf{m}_j$ indicates if $\tau_j$ belongs to the subtree rooted at $\tau_i$. It is then convenient to extract the desired subtree from the original $\mathbf{I}$ and $\mathbf{P}$ with $\mathbf{m}$. The new mask matrix is also generated by applying $\mathbf{m}$ to both rows and columns of the original $\mathbf{M}$. We present an example of to-subtree pruning in Figure~\ref{fig:tree_updating}(b).

\subsection{Pipelined Inference Framework}

\begin{figure}[t]
  \centering
  \includegraphics[width=0.9\columnwidth]{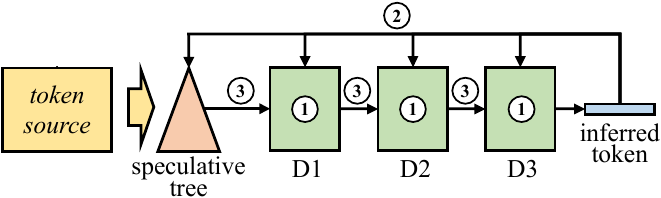}
  \caption{The workflow of \ours{}. D1, D2, and D3 are pipelined devices for inference.}
  \label{fig:pipeline_workflow}
  \small
\end{figure}

With the dynamic speculative tree presented in Section~\ref{sec:dst}, the pipelined inference framework of \ours{} works step by step. A pipeline step starts with 1) {\it node-wise computation}.
Following the inferred token at the last device, 2) {\it pruning propagation} is carried out.
The last stage of a pipeline step is 3) {\it inter-node transmission}. Meanwhile, new speculative tokens are inserted into the tree dynamically from a token source. The complete workflow of \ours{} is presented in Figure~\ref{fig:pipeline_workflow}. 

\subsubsection{Token source}
\label{sec:token_source}
Speculative tokens can be generated from either draft models~\cite{miao2024specinfer} or prefilled and inferred tokens~\cite{fu2024break}. However, the short prediction window lowers accuracy. In \ours{}, we do not have to generate a complete speculative token tree before inference but guarantee that a tree level is available at each pipeline step. Since the time of speculating tokens is relaxed, it is possible to trade off more computation power for higher prediction accuracy.


\begin{figure}[t]
  \centering
  \includegraphics[width=\columnwidth]{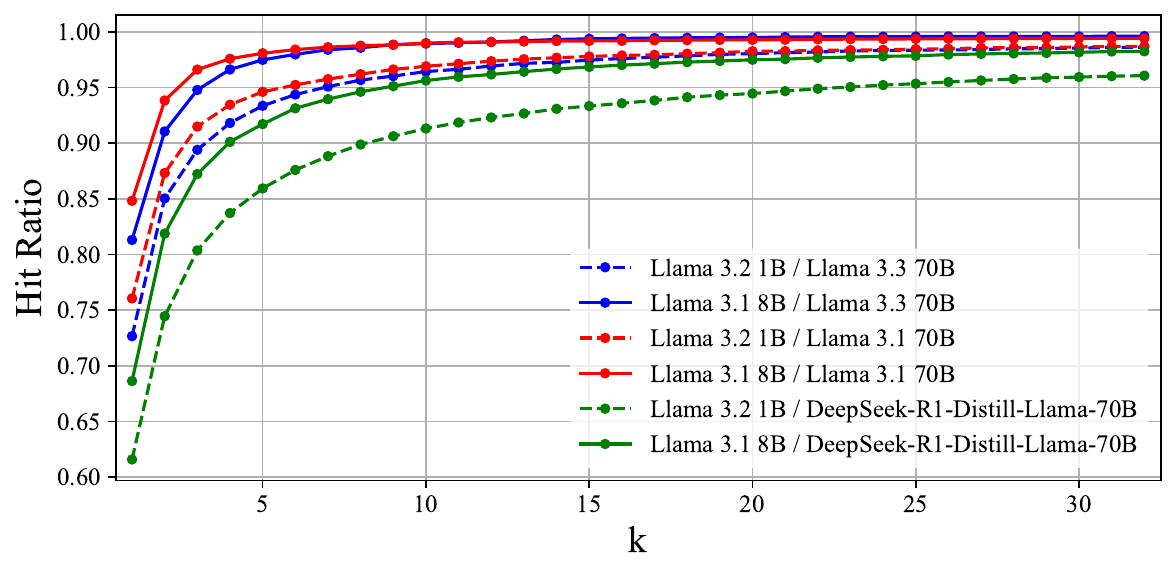}
  \caption{The accuracy of predicting a 70B model using top-$k$ output of a draft model. We evaluate various model pairs with literary comprehension tasks using ten novels from the {\it Project Gutenberg} (see appendix~\ref{sec:appendix-gutenberg} for details). Then the context length is around 20K tokens. }
  \label{fig:70Bvs8B}
  \small
\end{figure}

We carry out an experiment to evaluate the impact of using a draft model at different scales. We set up a couple of pairs of draft models and inference models, where the output of a 1B/8B model is used to predict the output of a 70B model. When evaluating a pair of models, the greedy strategy is used to select the next token of the 70B model, and we collect top-$k$ tokens from the draft model output. The experimental results are shown in Figure~\ref{fig:70Bvs8B}, where the x-axis represents increasing $k$ and the y-axis is the hit ratio. 
We observe that when using an 8B Llama3.1 draft model, the top-1 prediction accuracy reaches 0.81 and 0.85 when matching the outputs of Llama3.1-70B and Llama3.3-70B, respectively.
If we collect top-8 tokens, the accuracy reaches 0.99 and 0.99. Even using a 1B draft model, the accuracy also reaches 0.99 and 0.99 by collecting top-32 tokens, respectively. We also evaluate that using the 1B/8B draft model to predict a 70B Llama model that is supervised fine-tuned using distilled knowledge from DeepSeek-R1~\cite{guo2025deepseek} where the output contains chain-of-thought reasoning. As the output pattern differs, the prediction accuracy is lower but still reaches 0.96 and 0.98 when $k$ is 32. As for the inference time, the average TBT of the 1B/8B model on an NVIDIA L40 GPU is 17ms/67ms, and deploying the 70B models on 4 pipelined L40 GPUs takes 374ms to infer a token.

Thus, incorporating a larger draft model and using its top-$k$ output tokens to grow the speculative token tree helps \ours{} reach a high prediction accuracy without introducing additional complexity. Other techniques of generating speculative tokens are also compatible to \ours{}. 

\subsubsection{Pipeline-step stages}

The first stage is {\it node-wise computation}. At the beginning of each step, every device computes on the prepared token embeddings, while the last device infers the next token; the selection strictly follows the output of the large model, guaranteeing that its decoding performance is not degraded. Since multiple speculative sequences are processed in parallel, self-attention is computed with the tree mask~\cite{miao2024specinfer} rather than the conventional causal mask. The tree mask corresponds to the matrix $\mathbf{M}$ introduced in Section~\ref{sec:dst}. Whenever a new level of tree nodes is fed to the first device, a snapshot of the corresponding portion of $\mathbf{M}$ is also captured and transmitted, ensuring the tree mask is updated consistently. 


\begin{figure}[t]
  \centering
  \includegraphics[width=0.95\columnwidth]{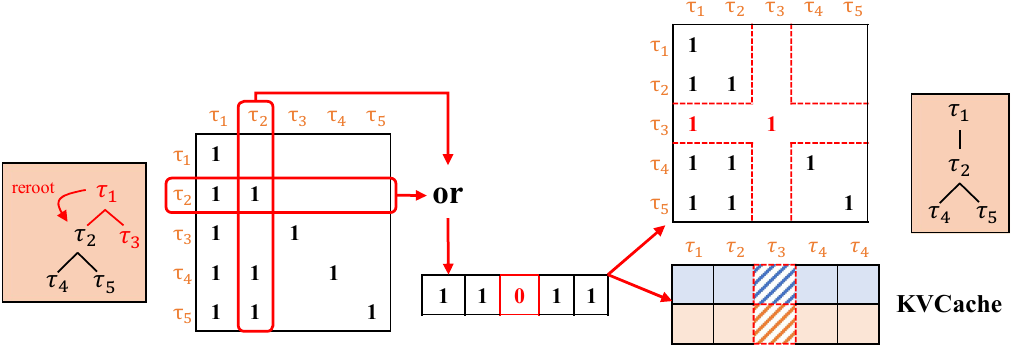}
  \caption{Updating the tree mask and \gls{kvcache} on a device.}
  \label{fig:treemask_kvcache_pruning}
  \small
\end{figure}

The second stage is {\it pruning propagation}. Once a new token is verified, the speculative tree is updated and pruned. Correspondingly, the tree mask, token embeddings, and the associated \gls{kvcache} are pruned across devices. This differs from the tree pruning in Section~\ref{sec:dst}, as the verified token and its \gls{kvcache} must be preserved. Figure~\ref{fig:treemask_kvcache_pruning} illustrates device-level pruning: the column and row corresponding to the rerooted token are extracted, combined with an {\tt OR} operation to form a temporary mask, and then applied to prune both the tree mask matrix and the \gls{kvcache}. 

\begin{algorithm}[t]
\caption{Pipeline-Step Stages}
\label{alg:pipeline_stages}
\begin{algorithmic}[1]
\Require a speculative tree $(C,P,M)$, number of devices $m$, device states $\{E_i, M_i, KV_i\}_{i=1}^m$ ($E$ is the embedding and $KV$ is the \gls{kvcache}), and a new speculative layer $L$
\medskip
\Statex \textbf{Node-wise computation}
\For{each layer in device $i$}
  \State $E_i \gets \text{layer}(E_i, M_i, KV_i)$
\EndFor
\State verify new token $\hat{\tau}$ on device $m$
\medskip
\Statex \textbf{Pruning propagation}
\State prune $M_i$, $E_i$, $L$ using mask $\mathrm{col}_{M_i}(\hat{\tau})$
\State prune $KV_i$ using mask $\mathrm{row}_{M_i}(\hat{\tau}) \;\lor\; \mathrm{col}_{M_i}(\hat{\tau})$
\State extract subtree rooted at $\hat{\tau}$ from $(C,P,M)$
\If{$L = \varnothing$} \Comment{Pipeline stall}
  \State $L \gets \{\hat{\tau}\}$
\EndIf
\medskip
\Statex \textbf{Inter‑node transmission}
\If{$i \neq m$}
  \State send remaining embeddings from device $i$ to $i+1$
\EndIf
\State send a speculative layer $L$ to device 1
\end{algorithmic}
\end{algorithm}

The last stage is inter-node transmission. As irrelevant tokens in the tree and token embeddings in devices have been pruned, the inter-node transmissions then occur. Every device sends embeddings that are computed and not pruned in this step to the next device, while the first device accepts a new level from the speculative tree. We present the process of different stages in Algorithm~\ref{alg:pipeline_stages}.

After the prefilling stage or a pipeline stall occurs, we have to refill the pipeline, which may involve only part of the stages or the entire pipeline depending on the situation. In the worst case, this refill follows the full-pipeline workflow, causing latency to degrade to the level of standard pipeline parallelism.

\subsubsection{Optimizing pipelined inference}
\label{sec:optimizations}

In a pipelined system, the performance is determined by the longest step. Thus, we need 1) to balance the workload to pipeline steps, and 2) to minimize workloads on individual devices. 

\noindent{\bf Fixed-width tree}.
If we grow the speculative tree exponentially, e.g., grow an $n$-ary tree, the workloads become highly skewed across devices. In such a case, a device closer to the pipeline end handles a lighter workload. To balance the workloads of devices, we propose to regulate the speculative tree to be ``fixed-width''. When expanding the tree, we follow a beam search-like rule: For each token at the tree bottom, we infer tokens at the next position with the draft model, and select $k$ of them with the highest probabilities. Then, we calculate cumulative probabilities of these tokens by multiplying probabilities of their ancestor nodes stored in $\mathbf{P}$. Among these tokens, we append $w$ of them with the highest cumulative probabilities to the tree. As a result, this constrains each level to $\leq w$ nodes. 

\begin{figure}[t]
  \centering
  \includegraphics[width=1\columnwidth]{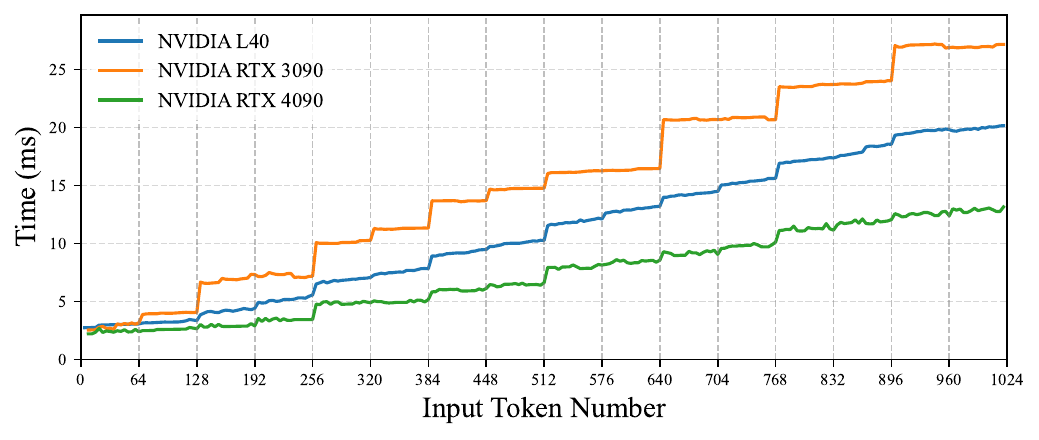}
  \caption{The inference time of a transformer layer on different GPUs. We vary the input token number while the context length is fixed to 2048.}
  \label{fig:scalinginference}
  \small
\end{figure}

\noindent{\bf Minimizing pipeline steps}.
\ours{} is expected to generate one token per step, so minimizing the cost of pipeline steps improves inference performance. Determining the number of input tokens per step becomes subtle because reducing the speculative tokens computed on a device shortens the execution time of a pipeline step but increases the risk of inferring an unseen token in the final stage, which may leave some stages idle, thereby impairing inference latency. Given a tree width $w$, the expectation of TBT can be calculated as
\begin{equation}
  \label{eq:expt_tbt}
  E(\mathcal{T}) = \max_{1 \leq i \leq m} t_i(w) + (1-P(w)) \cdot \sum_{i=1}^mt_i(w),
\end{equation}
where $m$ is the number of pipeline steps, $t_i(\cdot)$ is the time cost of step $i$, and $P(\cdot)$ is the probability of correct prediction at the last step. The first term corresponds to the time required to infer a token through the pipeline, regardless of correctness. The second term reflects the worst case: if the prediction fails, the entire pipeline becomes empty and must be refilled with a new speculative tree.  
Assuming the inference workload is evenly distributed across pipeline steps, Equation~\ref{eq:expt_tbt} can be simplified to
\begin{equation}
  \label{eq:expt_tbt2}
  E(\mathcal{T}) = t(w) + (1-P(w)) \cdot m \cdot t(w).
\end{equation}
If we use a draft model as discussed in Section~\ref{sec:token_source}, $P(w)$ would reach a plateau after tens of speculative tokens. This means that if more tokens are input, the risk of inferring an unseen token will not decrease. For $t(w)$, we experimentally measure the inference latency of a transformer layer on different devices, and present the results in Figure~\ref{fig:scalinginference}. The x-axis is the number of input tokens and the y-axis is the computation time in milliseconds. It is clear that the inference time linearly increases as more tokens are input. But instead of a smooth increase, the growth is flat and leaps every 64 tokens. This phenomenon arises because GPU matrix operations are executed in batches, where smaller fragments are padded during computation. From these observations, we therefore choose $w$ to saturate accuracy while avoiding the 64-token step changes.

\subsection{Other Design Choices}

\noindent{\bf Input more than one tree-level}.
We adopt a single-level-per-step strategy, which sustains a high prediction hit rate and is effective for typical pipelines. For very long pipelines (e.g., over 16 stages), multiple stages are grouped, and mini-batch concurrency can be applied within each group to balance prediction hit rate and latency while improving utilization.
We also evaluate feeding $l$ tree-levels per step, with an added token verification stage for the next $l{-}1$ levels. This approach showed worse performance: deeper trees sharply reduce hit rates, leading to frequent stalls, and multi-level verification and pruning introduce substantial GPU–CPU synchronization overhead.
A more adaptive design that selects prediction depth based on request characteristics could improve throughput, but would require complex optimization strategies, which we leave for future work.

\noindent{\bf Independent vs. Coupled Draft-Model Deployment}.
In our design, the draft model runs on a dedicated GPU. Many existing speculative decoding implementations instead integrate the draft model with the main model via tensor parallelism or place it on a pipeline-parallel GPU using leftover memory. Our approach avoids this for two reasons.
First, all GPUs must compute in parallel within each pipeline step to minimize latency. Co-locating the draft and main models on the same device would degrade performance~\cite{zhang2025swiftspec}. Second, the dedicated GPU hosting the draft model remains lightly loaded in both memory and compute, without competing for resources with the main inference tasks. 
So, we deploy the draft model on a separate GPU to maximize parallelism and minimize inference latency.

\noindent{\bf Lossless and Lossy Acceleration.}
\ours{} employs a lossless speculative decoding strategy, where \emph{lossless} means that every final token is produced by the large model. When the prediction from the draft model differs from that of the large model, the draft token is discarded. In practice, this losslessness is \emph{relative}. The accelerated model matches the non-accelerated model under greedy decoding only in specific settings—most reliably when using \texttt{fp32} precision or when GPU kernel accumulations follow a fixed deterministic order.
For example, two tokens can have nearly identical probabilities. With reduced-precision formats such as \texttt{bf16}, limited numerical accuracy and non-deterministic accumulation in fast kernels may cause their rankings to swap. A single mismatch can then cascade, altering nearly all subsequent tokens. Although this technically breaks exact greedy equivalence, the effect on overall performance is negligible.

Our framework can also adopt lossy verification strategies that balance output fidelity and latency depending on application requirements. We consider two representative options:  
(1) \textit{Relaxed verification without pipeline stall}: from the candidate tokens, select the one with the highest probability under the large model. This guarantees a $100\%$ hit rate and deterministic latency when pipeline flushing is costly.  
(2) \textit{Strict verification inspired by sampling}: accept a draft token if it lies within the top-$p$ cumulative probability set of the large model, and within that set choose the token with the highest draft-model probability. This improves the hit rate while incurring negligible accuracy loss.  
In this work, we follow the common practice in speculative decoding by validating against the top-1 prediction of the large model to ensure fair comparison across methods. Future work will examine the applicability of alternative verification strategies in diverse deployment settings.

\noindent{\bf Use Other Draft Generation Algorithms.}
\ours{} treats the token source as an independent component, allowing any draft generation method to serve as input. To support fixed-width trees, we prefer approaches with quantifiable prediction accuracy. Besides a dedicated draft model, low-cost options such as $n$-gram or database matching can be used, though their applicability is limited in tasks like translation or creative writing where contextual repetition is rare. Another direction is to integrate in-model prediction techniques~\cite{li2025eagle, cai2024medusa}, which provide high-accuracy drafts after targeted training and can be directly incorporated into our framework due to their independence from the computation of the main model. Finally, since different methods exhibit complementary strengths, constructing hybrid prediction trees may further improve accuracy and generality, which we leave for future work.






\section{Evaluation}
\label{sec:evaluation}

\noindent{\bf Implementation}.
We implement \ours{} based on PyTorch~\cite{imambi2021pytorch}, reusing  components from Hugging Face Transformers~\cite{wolf-etal-2020-transformers}. The framework provides operations for managing the speculative tree structure and its associated \gls{kvcache}. Our tree-attention masks are implemented by modifying the kernels in FlashInfer~\cite{ye2025flashinfer} for compatibility with the speculative decoding workflow. Inter-node communication is handled using NCCL~\cite{nccl}, and Redis~\cite{redis} serves as the synchronization mechanism across devices, with Lua scripts managing the dependency logic. We further apply asynchronous communication optimization, allowing data transfer to overlap with part of the pruning process, thereby reducing its impact on end-to-end latency.

For the multi-request version \oursdb{}, we implement dynamic batching using a custom ragged tensor structure. Prediction trees and token embeddings from different batches are packed into a single tensor, while a dedicated tensor is designed to record per-batch offsets for indexing and management. \gls{kvcache} storage adopts the PagedAttention~\cite{kwon2023efficient} approach.
In the prefilling stage, we follow the strategy in vLLM~\cite{kwon2023efficient}, which prioritizes new requests for minimizing first-token latency.

\noindent{\bf Testbed}.
We deploy \ours{} in a cluster of three servers interconnected via a 10 GbE network. Each server is equipped with 512 GB of main memory. For inference tasks, each pipeline step runs on a single GPU. One server contains four NVIDIA L40 GPUs (48 GB each), another contains four RTX 4090 GPUs (24 GB each), and the third contains eight RTX 3090 GPUs (24 GB each). Without NVLink, communication within a server uses PCIe and communication between servers uses Ethernet.

We select instruction-tuned versions of models from the Llama family~\cite{dubey2024llama} for evaluation. Specifically, we use Llama3.1-70B as the inference model and Llama3.2-1B as the draft model. The 70B inference model consists of 80 transformer layers and 2 linear layers. 
The default {\tt bfloat16} precision is used for all computations.
To comprehensively evaluate \ours{}, we build two 8-stage pipelines with a single-server and a two-server configuration, and we always use an individual L40 GPU running the draft model.

\textit{The single-server configuration}: It consists of 8 RTX 3090 GPUs, where 6 GPUs host 10 transformer layers each, and the remaining 2 GPUs host 10 transformer layers plus a linear layer each. On each GPU, approximately 20 GB of device memory is allocated for model parameters and activations, with the remaining memory dedicated to the \gls{kvcache}.

{\it The two-server configuration}: It consists of 4 RTX 3090 GPUs in a server and 4 RTX 4090 in another. The layer setting is the same as the single-server configuration. 


\begin{figure}[t]
  \centering
  \includegraphics[width=\columnwidth]{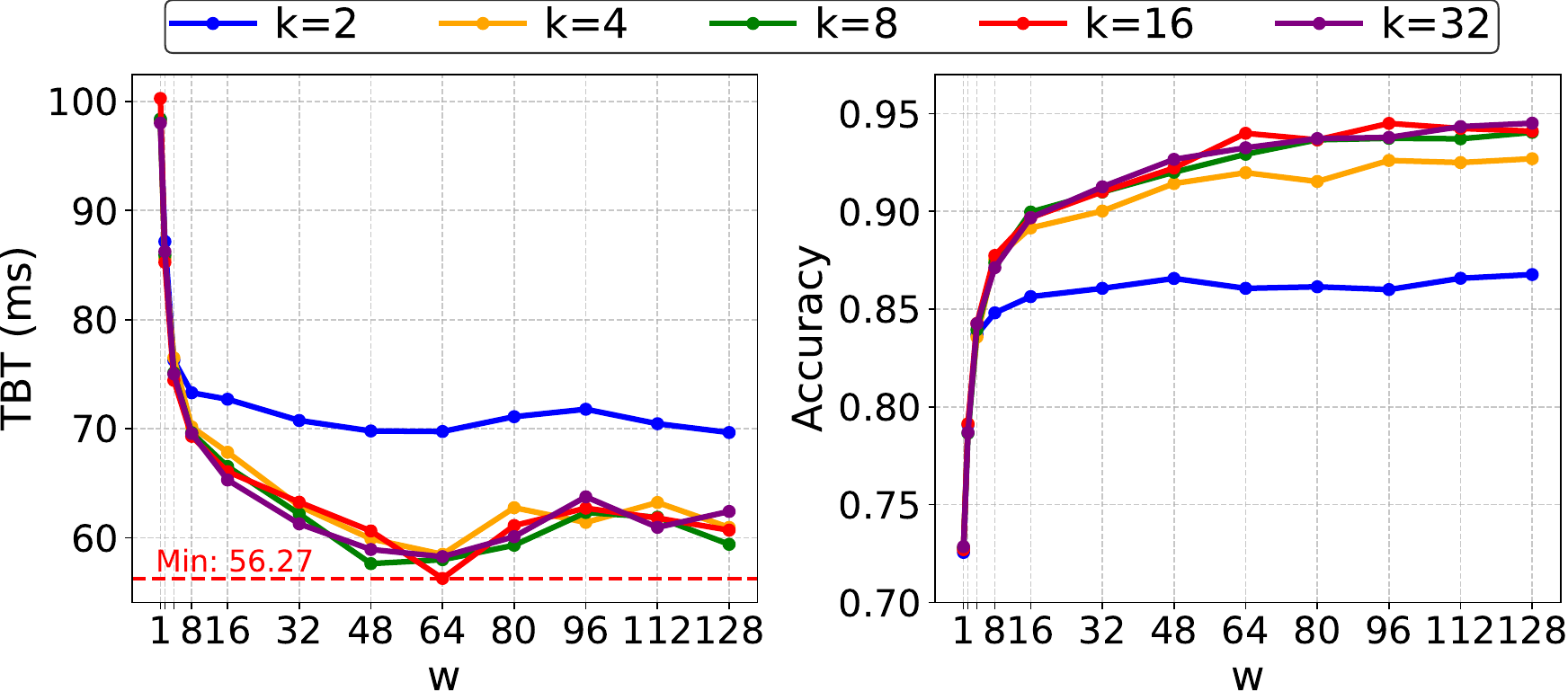}
  \caption{Average TBT and accuracy when varying $w$ and $k$}
  \label{fig:exp-TreePara-MIX}
  \small
\end{figure}

\noindent{\bf Baselines}.
The {\it standard pipeline parallelism (PP)} processes tokens autoregressively for each batch without any speculative mechanisms.

The second baseline is \textit{SpecInfer}~\cite{miao2024specinfer}, the state-of-the-art in tree-based speculative decoding. Its open-source implementation cannot run stably under multi-GPU pipeline parallelism, frequently resulting in out-of-memory or runtime errors due to a hard-coded context length limitation (Appendix~\ref{sec:appendix-specinfer}). To ensure fairness and eliminate framework-level effects, we re-implement SpecInfer within our framework using the same Llama3.2-1B draft model as \ours{}. Both methods adopt the asynchronous scheduling optimization, which overlaps draft model computation with pipeline stages to reduce the latency overhead of the larger draft model. For the prediction tree, we follow the original setting in~\cite{miao2024specinfer}, using an 8-level structure with branching $\langle 1, 1, 3, 1, 1, 1, 1, 1 \rangle$, where the wider third level enables limited exploration while keeping the number of speculative tokens manageable.

The third baseline, {\it EAGLE3}~\cite{li2025eagle}, augments inference with additional pre-trained parameters to predict subsequent tokens and enable parallel verification. We use their open-source implementation and pre-trained weights, adopting Llama3.3-70B, whose architecture and computational cost are identical to those of Llama3.1-70B (Appendix~\ref{sec:appendix-eagle3}), as the inference model under pipeline parallelism. EAGLE3 cannot run on a two-server configuration.

The fourth baseline is pipeline parallelism implemented in vLLM~\cite{kwon2023efficient} (\textit{PP-vLLM} and \textit{TP/PP-vLLM}), a widely-adopted inference framework with comprehensive optimizations supporting diverse models. \textit{TP/PP-vLLM} refers to a configuration where tensor parallelism runs on four GPUs within a single server, while pipeline parallelism spans two servers. We leverage its pipeline parallelism implementation to execute Llama3.1-70B inference. Notably, as of the current version (0.8.3), vLLM does not support speculative decoding in combination with pipeline parallelism.


\noindent{\bf Benchmarks}.
We evaluate \ours{} and baselines with a wide range of benchmarks, including HumanEval-X~\cite{zheng2023codegeex,chen2021evaluating} for programming, DROP~\cite{Dua2019DROP} for reading comprehension, MMLU~\cite{hendryckstest2021, hendrycks2021ethics} for general question answering, WMT14~\cite{bojar-EtAl:2014:W14-33} for translation, TriviaQA-Wiki~\cite{2017arXivtriviaqa} for knowledge reasoning, and GSM8K~\cite{cobbe2021gsm8k} for mathematics. We randomly select 20 samples from each benchmark (120 in total) for evaluation.

\subsection{Selection of Prediction Tree Parameters}

\begin{figure}[t]
  \centering
  \includegraphics[width=\columnwidth]{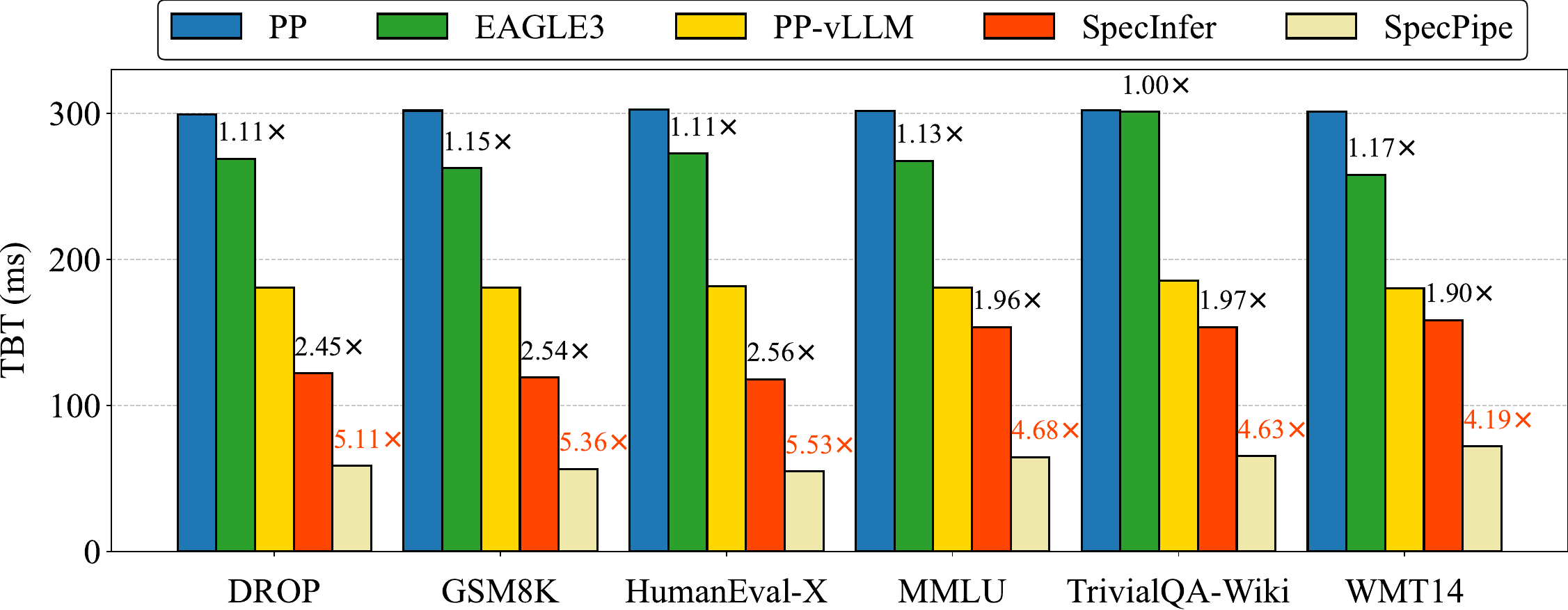}
  \caption{The average TBT of different inference methods using pipeline parallelism (single-server configuration).}
  \label{fig:exp-Analytics-Latency-single}
  \small
\end{figure}

We first evaluate how the tree width $w$ and the candidate child number $k$ affect inference performance. When expanding a speculative tree, $k$ candidates are generated for each token at the tree bottom. The cumulative probabilities of these candidates are computed, and up to $w$ candidates with the highest probabilities are appended to the tree. Experiments are conducted in the one-server configuration with $w \in \{1, 2, 4, 8, 16, 32, 48, 64, 80, 112, 128\}$ and $k \in \{2, 4, 8, 16, 32\}$. At each step, one tree level is fed to the pipeline, and we measure both the \gls{tbt} (in milliseconds) and the accuracy, where accuracy is defined as the proportion of speculative tree layers containing the ground-truth token from the inference model. Results are averaged over all datasets.

Figure~\ref{fig:exp-TreePara-MIX} shows that increasing tree width improves accuracy, as more candidates are available to match the real token. However, the gain diminishes quickly because most ground-truth tokens already rank highly in the candidate list. A larger width also reduces the risk of pipeline stalls, initially lowering latency, but once accuracy saturates, the added computation per step causes latency to rise (Section~\ref{sec:optimizations}). A small $k$ lowers accuracy because the most probable next-layer tokens may all originate from the same parent. With $k=2$, accuracy drops sharply, leading to frequent stalls and higher latency. When $k \geq 4$, accuracy remains stable, and the effect on latency becomes negligible. Based on these observations, we set $w=64$ and $k=16$ for all subsequent experiments.

\begin{figure*}[t]
  \centering
  \includegraphics[width=\textwidth]{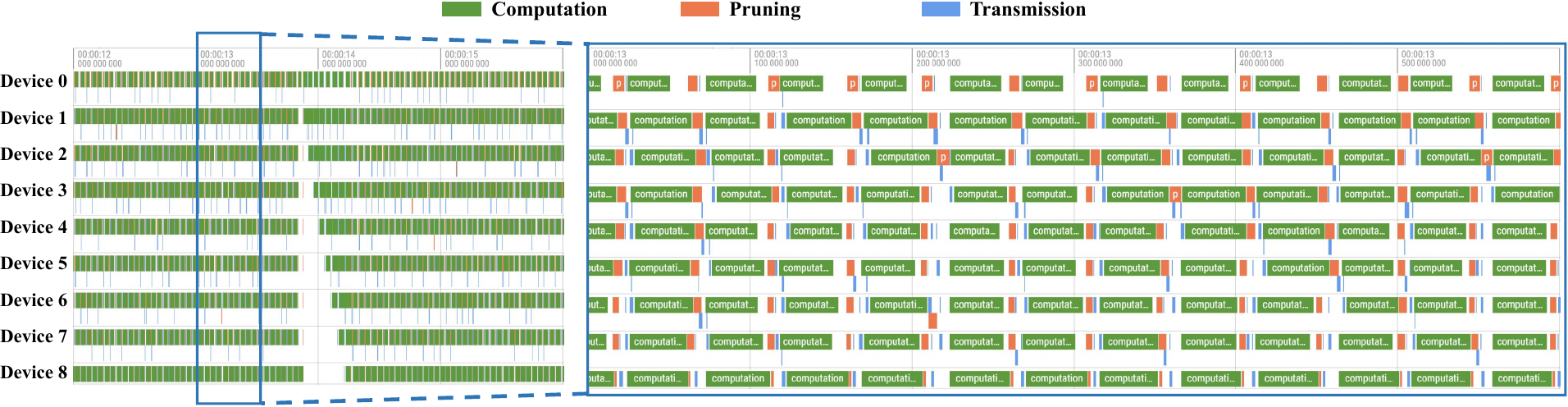}
  \caption{Profiling of \ours{} on the one-server configuration while inferring a request from HumanEval-x. ({\bf Left}) A zoom-out figure of 4-second duration (12,000–16,000 ms). Device 0 holds the draft model, and Device 1-8 hold the inference model. ({\bf Right}) A zoom-in figure of 500 ms interval (13,000–13,500 ms).}
  \label{fig:exp-pipeline}
\end{figure*}

\subsection{Single-request Inference}
We evaluate \ours{} and the baselines using the one-batch setup. In this experiment, a single batch is sent to the pipeline at a time for all methods, and the next batch is issued only after the previous one completes. We measure the TBT in milliseconds. For pipeline parallelism (PP and PP-vLLM) as well as PP with various speculative decoding methods (SpecInfer and EAGLE3).

The single-server results are shown in Figure~\ref{fig:exp-Analytics-Latency-single}, where the x-axis represents different benchmarks and the y-axis is the average TBT. \ours{} significantly reduces inference latency. For example, in HumanEval-X, the average TBTs of PP and PP-vLLM are 302.39~ms and 181.41~ms, respectively. With speculative decoding, the TBT becomes 272.30~ms (EAGLE3) and 117.92~ms (SpecInfer). \ours{} further reduces the average TBT to 54.67~ms. Notably, EAGLE3 has higher latency than PP-vLLM because it predicts tokens using a few tuned layers of the inference model, requiring a full pipeline pass to generate speculative tokens, which largely offsets the benefits of speculative decoding. Similar performance trends are observed in other benchmarks. Overall, \ours{} improves TBT by 4.19$\times$--5.53$\times$ over PP, 2.50$\times$--3.32$\times$ over PP-vLLM, 3.58$\times$--4.98$\times$ over EAGLE3, and 2.09$\times$--2.38$\times$ over SpecInfer. 
\begin{figure}[t]
  \centering
  \includegraphics[width=\columnwidth]{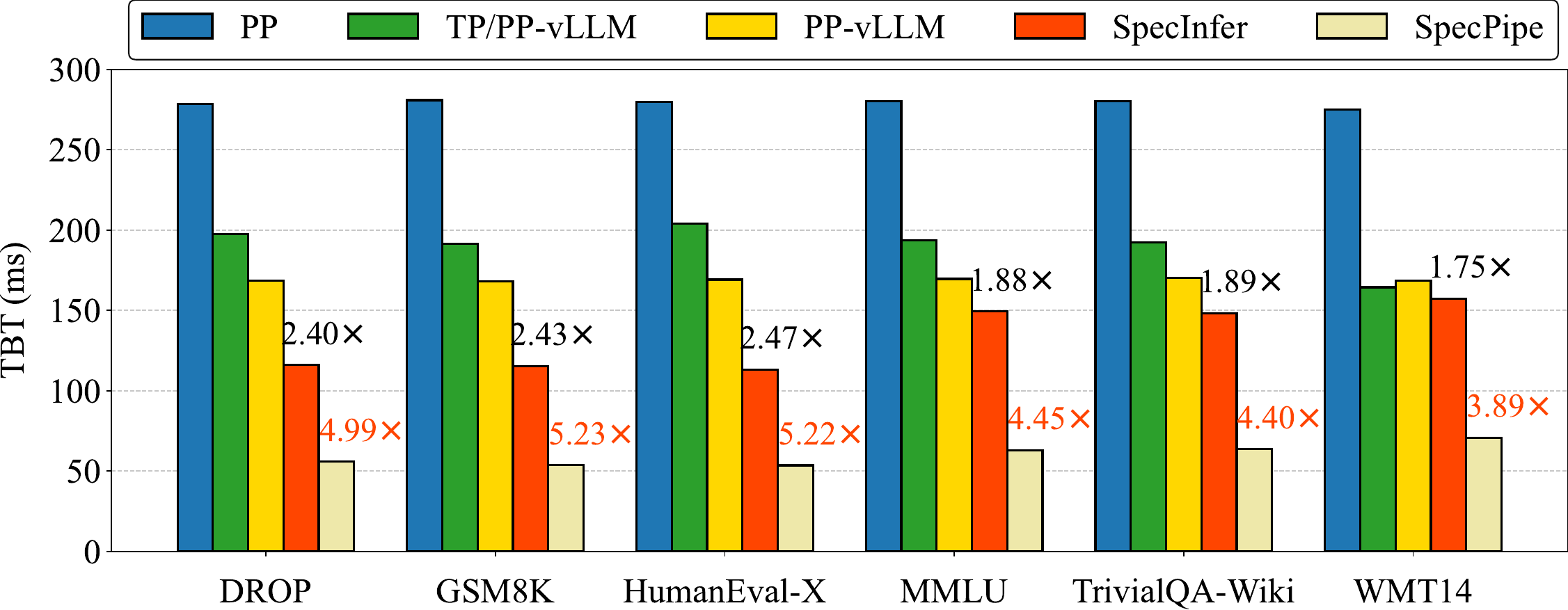}
  \caption{The average TBT of different inference methods using pipeline parallelism (two-server configuration).}
  \label{fig:exp-Analytics-Latency-two}
  \small
\end{figure}

We visualize one of inference requests executed by \ours{} in Figure~\ref{fig:exp-pipeline}, where the y-axis is different devices and the x-axis is the elapsing time. We mark computation, pruning, and transmission with different color. In the zoom-out figure at the left side, a clear fissure representing a pipeline flush due to misprediction is observable. In the zoom-in figure at the right side, we can seen that the pipeline steps are well aligned, and most time is spent on computation.

This setup also allows single-node tensor parallelism, which offers lower latency than any pipeline-based method. Our study, however, targets accelerating pipeline parallelism with \ours{}. We retain this configuration to evaluate EAGLE3, which is limited to a single node, ensuring fairness. All methods operate independently of the high PCIe bandwidth available in this environment. With speculative tree embeddings, inter-node traffic rises from 16 KiB to at most 1 MiB (a $64\times$ increase). \ours{} mitigates this by overlapping pruning and transfer through asynchronous communication; since pruning takes about \textasciitilde 5 ms, even a 10 GbE network with millisecond-level latency suffices.
Figure~\ref{fig:exp-Analytics-Latency-two} reports the average TBT of different methods on the two-server with 10 GbE network configuration. \ours{} consistently achieves the best TBT among all baselines, improving it by 3.89$\times$--5.23$\times$ over PP, 2.38$\times$--3.16$\times$ over PP-vLLM, 2.55$\times$--3.44$\times$ over TP/PP-vLLM, and 2.08$\times$--2.37$\times$ over SpecInfer. These results further indicate that our system is minimally affected by inter-node communication bandwidth.



\begin{figure}[t]
  \centering
  \includegraphics[width=\columnwidth]{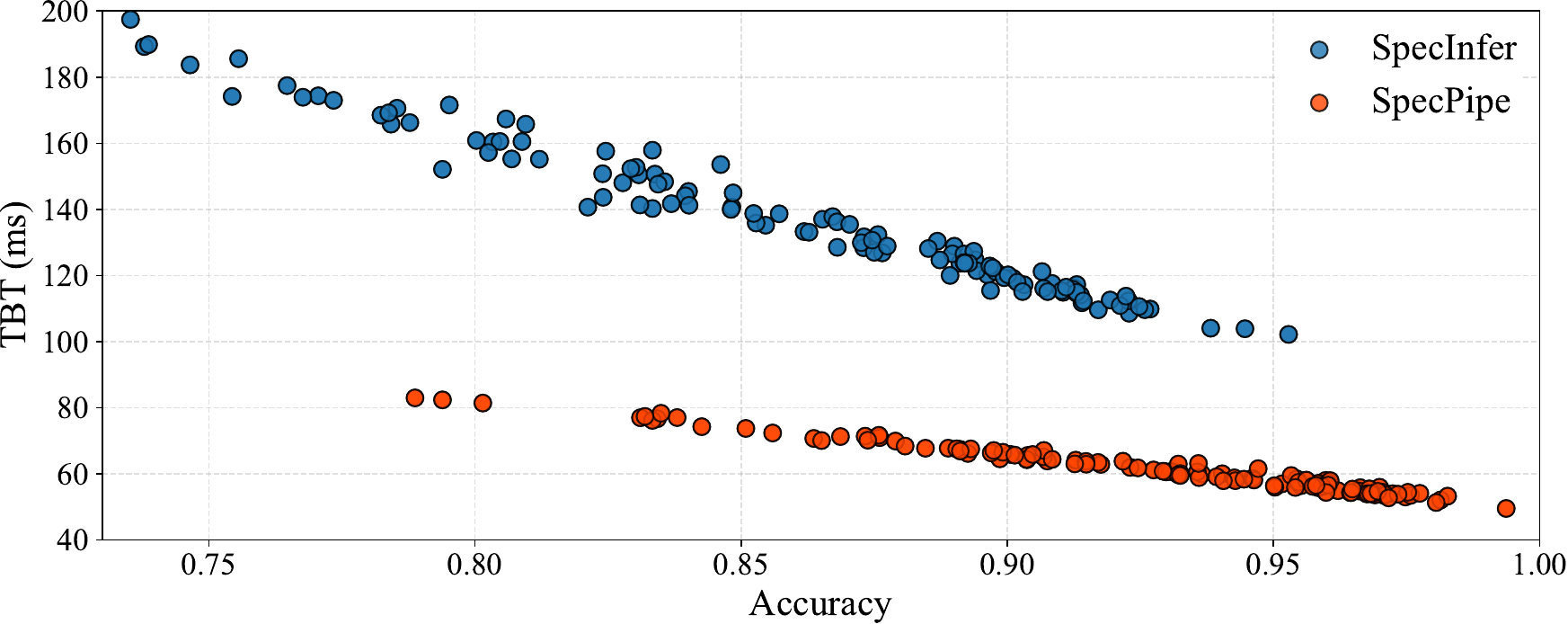}
  \caption{Scatter plot of prediction tree hit rate versus \gls{tbt} for SpecPipe and SpecInfer.}
  \label{fig:exp-Analytics-ACC}
  \small
\end{figure}

To assess the latency impact of pipeline refills caused by prediction misses, we compare the static tree of SpecInfer with the dynamic tree in \ours{} in terms of prediction accuracy. Figure~\ref{fig:exp-Analytics-ACC} plots accuracy (x-axis) against TBT in milliseconds (y-axis). \ours{} achieves consistently higher accuracy, as pruning in the dynamic tree preserves more valid predictions under the same computation budget. By proactively supplementing potential break points, the dynamic tree overlaps part of the refill cost with ongoing correct predictions, thereby reducing the penalty of misses. Consequently, even at the same accuracy level, \ours{} attains lower latency than SpecInfer by effectively hiding draft computation overhead, confirming the efficiency of the proposed design.

\subsection{Stochastic Decoding}

\begin{figure}[t]
  \centering
  \includegraphics[width=\columnwidth]{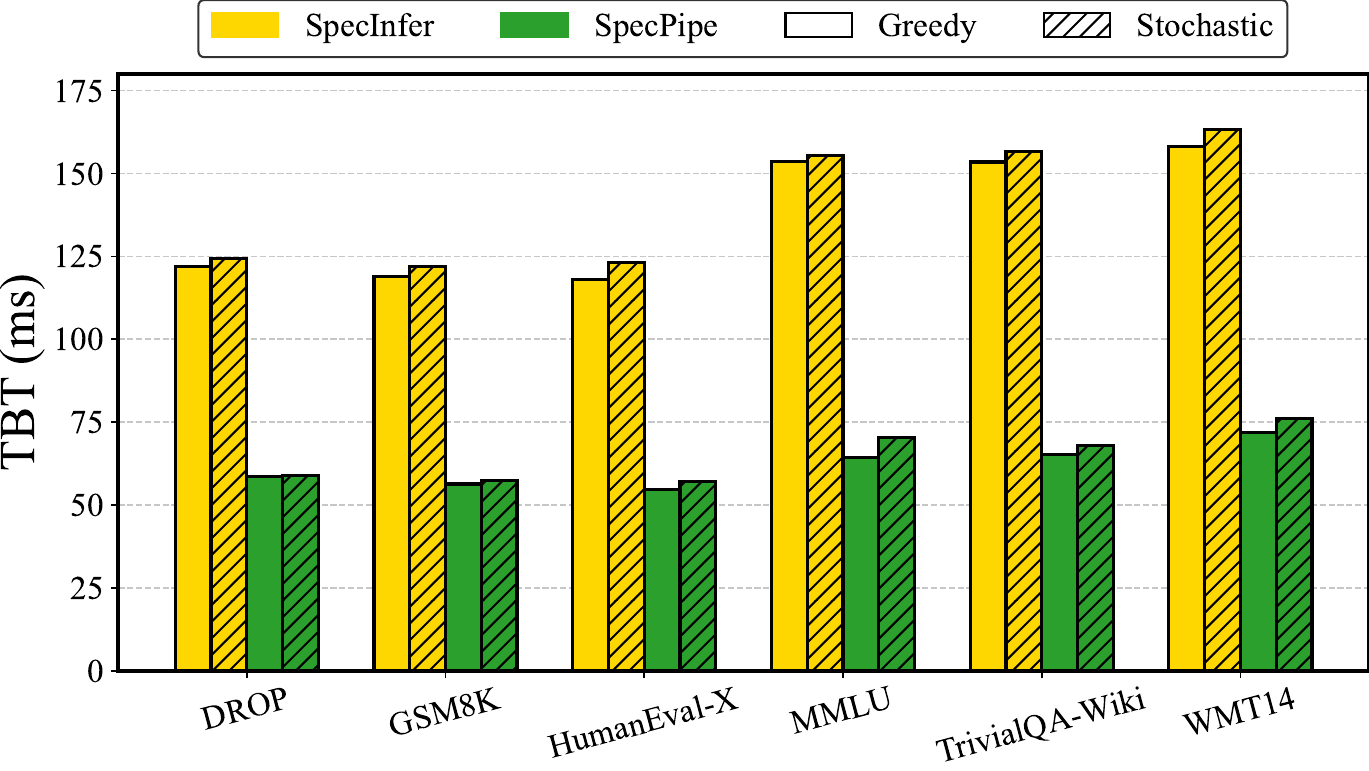}
  \caption{\gls{tbt} comparison between \ours{} and SpecInfer under greedy and stochastic decoding.}
  \label{fig:exp-Stochastic-MIX}
  \small
\end{figure}

To ensure output diversity, large language models often employ stochastic decoding. We therefore switch to stochastic decoding in our experiments to evaluate the stability of \ours{}. We compare \ours{} and SpecInfer under a single-server configuration with the following setup: temperature is set to 0.6, and sampling continues until either the cumulative probability reaches 0.9 or the number of candidates reaches 80. Due to the inherent randomness of stochastic decoding, each query is executed five times, and we report the average TBT and accuracy.  

The experimental results are presented in Figure~\ref{fig:exp-Stochastic-MIX}. We observe that switching from greedy decoding to stochastic selection leads to a slight increase in average latency for both SpecInfer and \ours. On average, latency increases by 2.54\% for SpecInfer and 4.39\% for \ours. This slight increase stems from a small reduction in prediction hit rate when using stochastic decoding. The results demonstrate that our method performs consistently under both decoding strategies.

\subsection{Multi-request Inference}

We evaluate the inference throughput of various methods using the multi-request setup. Ten queries are randomly selected from each dataset to form a pool of 60 queries. A naive multi-batch scheduling strategy is applied: given a batch size $B$, new tasks are launched whenever the number of concurrently running tasks is below $B$, ensuring that all requests continuously utilize the available computation resources. \ours{} and SpecInfer adopts an iteration-based scheduling strategy, \oursdb{} uses a dynamic batching approach, and vLLM combines dynamic batching with mini-batch scheduling.

\begin{figure}[t]
  \centering
  \includegraphics[width=\columnwidth]{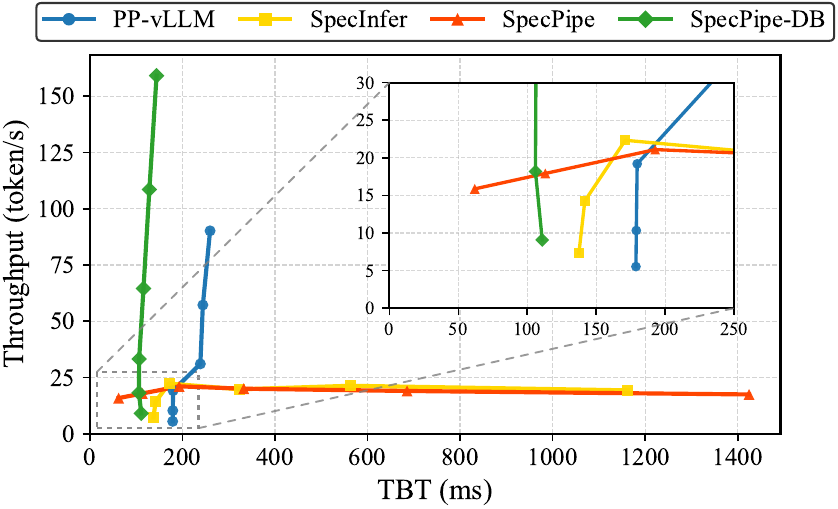}
  \caption{Throughput–TBT trade-off of different pipeline parallelism methods.}
  \label{fig:exp-Throughput-vs-TBT}
  \small
\end{figure}

We compare \ours{} and \oursdb{} with PP, SpecInfer, and PP-vLLM under batch sizes $B \in \{1,2,4,8,16,32\}$. For each setting, we measure inference throughput (tokens/s) and average single-request TBT, where each point on a curve corresponds to one batch size in $B$. The results are shown in Figure~\ref{fig:exp-Throughput-vs-TBT}. \ours{} achieves the lowest TBT in the single-batch case due to its minimal autoregressive delay. As batch size increases, PP-vLLM and SpecInfer gradually surpass \ours{} in throughput. In contrast, \oursdb{} sustains the highest throughput for $B \geq 2$, outperforming PP-vLLM by 1.64$\times$–2.08$\times$ in throughput while reducing TBT by 1.61$\times$–2.06$\times$. The higher TBT of \oursdb{} compared with \ours{} arises from the additional computation of dynamic batching. Profiling shows that LLM inference and batching overhead contribute roughly the same. Since the current ragged tensor implementation in PyTorch prevents automatic kernel fusion, one of future work is to design fused CUDA kernels for further optimizing batching overhead.

\begin{figure}[t]
  \centering
  \includegraphics[width=\columnwidth]{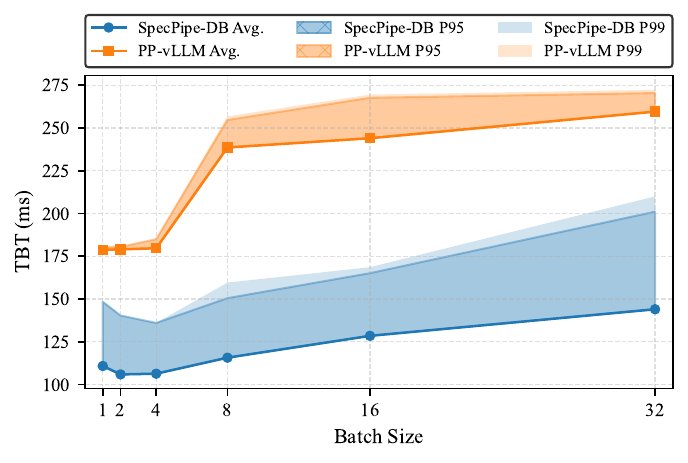}
  \caption{Tail latency comparison between SpecPipe-DB and PP-vLLM}
  \label{fig:exp-tail-latency}
  \small
\end{figure}



Figure~\ref{fig:exp-tail-latency} reports the tail latency ($Px$ percentile of TBT) of \oursdb{} under different batch sizes. Although its tail latency grows faster than the average, it consistently stays below that of vLLM, showing robust acceleration even in the worst case. The larger gap between P99 and average latency of \ours{} is due to occasional misprediction, which results in pipeline flush.

\subsection{GPU Memory Analysis of \ours{}}

We evaluate memory usage of Llama-3.1-70B with hidden size $d=8192$, $L=80$ layers, $n_{\text{heads}}=64$, grouped-query attention with $n_{\text{kv}}=8$, head dimension $d_h=128$, and feed-forward size $f=28672$. Parameters are stored in bfloat16. The model is partitioned across 8 GPUs with 10 layers per shard. GPU 0 additionally hosts the embedding layer and GPU 7 the output projection. The vocabulary size is $V=128{,}256$. A fixed context budget of $81{,}920$ tokens is reserved as the \gls{kvcache} for all requests, which is shared by both the speculative tree nodes and the tokens generated by the large model. The parameter count per transformer layer is \(N_{\text{layer}} = 4d^2 + 3df\).

\begin{table}[h]
\centering
\caption{Per-GPU memory usage under pipeline parallelism. Units are GiB ($2^{30}$ bytes).}
\renewcommand{\arraystretch}{1.1}
\resizebox{\linewidth}{!}{%
\begin{tabular}{l l r}
\toprule
GPU & Components & Total (GiB) \\
\midrule
0 (with embedding) & 10 layers + embedding + KV + runtime & 23.5 \\
1–6 & 10 layers + KV + runtime & 21.5 \\
7 (with output) & 10 layers + output + KV + runtime & 23.5 \\
\bottomrule
\end{tabular}}

\label{tab:mem}
\end{table}

The speculative tree adds at most 4~MiB GPU memory per shard in the worst case, and pruning reduces this further. Runtime kernels are supported by a 256~MiB cache. Overall, memory demand is close to standard pipeline parallelism, with only the first and last GPUs approaching the 24~GiB limit.

\section{Related Work} 
\label{sec:RelatedWork}

\subsection{Speculative Decoding} 
Early work introduces block-parallel decoding, which accelerates generation through multi-step prediction but remains limited to greedy decoding~\cite{NEURIPS2018_c4127b91}. Speculative decoding then establishes a draft-verify paradigm~\cite{xia2024unlocking}, where a lightweight draft model proposes tokens and the inference model verifies them in parallel. This requires the draft model to be tweaked for aligning with the large inference model~\cite{leviathan2023fast,xia2023speculative}. These foundations motivate subsequent designs which further optimizes LLM inference with speculative decoding.

One line of research focuses on improving draft model quality. DistillSpec applies knowledge distillation to align the draft with the inference model~\cite{zhou2023distillspec},and the draft model is online updated for further adapting its distribution~\cite{liu2023online}. Medusa equips the inference model with auxiliary heads to propose multiple candidates~\cite{cai2024medusa}, the EAGLE series applies context-aware strategies to reduce uncertainty~\cite{li2024eagle,li2024eagle2,li2025eagle}, and BiLD combines a small draft model with rollback and fallback mechanisms~\cite{kim2024speculative}. Retrieval-based approaches such as REST replace the draft model with external retrieval~\cite{he2023rest}, while multi-draft selection methods like SpecTr improve candidate choice via optimal transport~\cite{sun2023spectr}. Although these methods improve acceptance rates, they often require fine-tuning, large-scale data processing, or architectural modifications, which raise engineering costs and undermine plug-and-play usability. Their scalability and stability across tasks depend heavily on the datasets used for training.

Another direction avoids training a separate draft model. Self-speculative decoding generates drafts by skipping intermediate layers of the inference model~\cite{zhang2023draft}, and lookahead decoding integrates the predicted tokens directly into the ongoing inference sequence~\cite{fu2024break}. LayerSkip~\cite{elhoushi2024layerskip} and SWIFT~\cite{xia2024swift} integrate early exits and dynamic layer skipping to further reduce overhead. These approaches ease deployment and eliminate training costs, but have to trade off prediction quality on complex tasks and lack systematic validation in large distributed environments. 

Tree-structured speculative decoding has the advantage of inferring the common ancestor of speculative tokens once. Staged speculative decoding employs a staged tree to increase accepted tokens in small-batch and single-device scenarios~\cite{spector2023accelerating}, while SpecInfer organizes candidate production and verification in a tree-based manner~\cite{miao2024specinfer}. Despite theoretical efficiency, both approaches are constrained by limited node budgets and concentrated verification, which cap accuracy improvements and limit scalability. System-level implementations such as TensorRT-LLM~\cite{TensorRT-LLM} and Ghidorah~\cite{wei2025ghidorah} integrate speculative decoding with optimized kernels or hetero-core execution, but they target single-node or edge deployments and overlook pipeline-parallel scaling. In contrast, our approach overlaps verification with communication to preserve efficiency across multi-node pipelines, fully exploiting pipeline parallelism for adaptive acceleration.

\begin{table}[h]
\caption{System-level comparison of speculative decoding methods. 
A check mark indicates support for the feature, while a cross mark indicates 
that the feature is not supported or has not been investigated.}
\centering
\setlength{\tabcolsep}{12pt} 
\begin{tabular}{lcccc}
\toprule
\textbf{Method} & \textbf{1} & \textbf{2} & \textbf{3} & \textbf{4} \\
\midrule
SpecPipe (Ours)   & \cmark & \cmark & \cmark & \cmark \\
SpecInfer~\cite{miao2024specinfer}         & \xmark & \cmark & \cmark & \xmark \\
EAGLE~\cite{li2024eagle,li2024eagle2,li2025eagle}             & \xmark & \xmark & \cmark & \cmark \\
Medusa~\cite{cai2024medusa}            & \xmark & \xmark & \xmark & \xmark \\
Self-Speculative~\cite{zhang2023draft}  & \xmark & \cmark & \xmark & \xmark \\
Lookahead~\cite{fu2024break}         & \cmark & \cmark & \xmark & \xmark \\
\bottomrule
\end{tabular}
\label{tab:system_comparison}
\begin{minipage}{\columnwidth}
\footnotesize
\(\\\)
\textbf{Feature index:}  \\
1 = Draft compute adds no extra latency,  \\
2 = No extra training,  \\
3 = Large batch supporting,  \\
4 = Dynamic draft.  
\end{minipage}
\end{table}

Alternative strategies explore further trade-offs. Reward-guided decoding integrates task-specific signals to balance fidelity and efficiency~\cite{liao2025reward}, but this relaxes strict distributional equivalence. Retrieval-driven methods depend on external data systems and incur additional latency, while hardware-oriented systems are confined to specific infrastructures. SpecPipe positions itself differently: it achieves training-free acceleration and scales robustly in pipeline-parallel environments. Table~\ref{tab:system_comparison} summarizes representative methods in terms of latency, training cost, scalability, and adaptability, highlighting the balanced advantages of our design.

\subsection{Accelerating Pipeline Parallelism in LLM Inference}

Research on accelerating pipeline parallelism during inference primarily falls into three categories. (1) Scheduling and batching methods, such as Orca~\cite{yu2022orca} and Sarathi-Serve~\cite{agrawal2024taming}, reduce pipeline bubbles by enabling iteration-level or stall-free batching. DistServe~\cite{zhong2024distserve} disaggregates prefill and decoding to balance resource use under latency constraints. (2) Memory management systems, most notably vLLM with PagedAttention~\cite{kwon2023efficient}, support fine-grained dynamic batching that improves utilization across pipeline stages. (3) Adaptive and multi-tenant strategies, including EE-LLM~\cite{chen2023ee} and AlpaServe~\cite{li2023alpaserve}, leverage early exit techniques or elastic parallelization to enhance efficiency in large-scale or multi-tenant deployments. In parallel, speculative methods such as PipeInfer~\cite{butler2024pipeinfer} and SPACE~\cite{yi2024generation} demonstrate how asynchronous speculation or semi-autoregressive decoding can further reduce latency. Together, these approaches demonstrate the effectiveness of pipeline parallelism in improving throughput and robustness across diverse workloads. However, there has been little progress on efficiently utilizing GPUs for small-batch inference, and achieving low-latency integration of pipeline parallelism with speculative decoding remains an open direction.

\section{Conclusion} 
\label{sec:ConclusionAndFutureWork}

In this paper, we present \ours{}, a speculative decoding framework designed to accelerate pipeline parallelism-based LLM inference. By filling the inference pipeline with speculative tokens of single-request step-by-step, \ours{} targets one token generation per pipeline step, significantly reducing TBT. This is achieved through a dynamically maintained speculative token tree that receives predictions from a large, high-accuracy draft model and feeds tree-level representations into the pipeline. Each pipeline step executes computation, pruning, and communication, with performance further improved by a fixed-width tree design that balances workloads and minimizes stalls.
Experiments show that \ours{} substantially reduces inference latency and improves inference throughput, outperforming both pipeline parallelism and prior speculative decoding methods. Its dynamic batching extension \oursdb{} further improves efficiency under multi-request workloads.

\newpage

\bibliographystyle{ACM-Reference-Format}
\bibliography{main}

\newpage

\appendix




\section{Experimental Fairness}
\label{sec:appendix-fairness}

\subsection{SpecInfer}
\label{sec:appendix-specinfer}
For completeness, we note that SpecInfer was evaluated using the \texttt{flexflow-serve} implementation referenced in~\cite{miao2024specinfer}. This system enforces a hard-coded token length limit (1024 tokens), and attempts to modify this parameter consistently triggered runtime errors. 
We observed persistent out-of-memory failures when running on 9 RTX 3090 GPUs. We therefore switched to a 10 RTX 3090 GPUs configuration and fixed \texttt{max\_seq\_length}=512 to maintain stability, which required truncating prompts to 800 tokens and left roughly 220 tokens for decoding. These constraints deviate from the intended evaluation protocol: the truncated inputs are largely incomplete and cannot faithfully represent the original datasets, and the input truncation further causes the model to fall into repetitive output loops that the framework fails to terminate. As a result, subsequent predictions are continuously accepted, making the measurements unreliable and not comparable to the main results; we report them here only for reproducibility and reference.

Table~\ref{tab:single_batch_latency} reports the single-batch latency on six datasets, and Table~\ref{tab:multi_batch_latency_throughput} summarizes latency and throughput under different batch sizes.

\begin{table}[ht]
\small
\setlength{\tabcolsep}{6pt}
\caption{Single-batch latency per dataset of SpecInfer.}
\centering
\begin{adjustbox}{max width=\linewidth}
\begin{tabular}{l r}
\toprule
Dataset & Latency (ms) \\
\midrule
DROP            & 111.43 \\
GSM8K           & 79.25 \\
HumanEval-X     & 71.46 \\
MMLU            & 119.08 \\
TrivialQA\text{-}Wiki & 65.66 \\
WMT14           & 82.92 \\
\bottomrule
\end{tabular}
\end{adjustbox}
\label{tab:single_batch_latency}
\end{table}

\begin{table}[h]
\small
\setlength{\tabcolsep}{6pt}
\caption{Latency and throughput under different batch sizes of SpecInfer.}
\centering
\begin{adjustbox}{max width=\linewidth}
\begin{tabular}{r r r}
\toprule
Batch size & Latency (ms) & Throughput (tokens/s) \\
\midrule
1   & 197.23 & 4.84 \\
2   & 154.88 & 6.15 \\
4   & 212.58 & 4.49 \\
8   & 98.05  & 9.73 \\
16  & 100.17 & 9.53 \\
32  & 217.67 & 4.39 \\
\bottomrule
\end{tabular}
\end{adjustbox}
\label{tab:multi_batch_latency_throughput}
\end{table}

\subsection{EAGLE3}
\label{sec:appendix-eagle3}
EAGLE3 provides an open-source, pretrained Llama-3.3-70B model. 
This model matches the Llama-3.1-70B used in our experiments in both parameter scale and architectural design. 
Although differences in training data may lead to variations in output content or sequence length, the per-token computational cost is determined solely by the model architecture and parameter count, which are identical. 
Therefore, evaluating the average \gls{tbt} of EAGLE3 offers a fair comparison with our system in terms of computational efficiency, even if output-level behaviors differ.



\section{Project Gutenberg Dataset}
\label{sec:appendix-gutenberg}
To evaluate the prediction hit rate of draft models, we assembled a test set of ten books from Project Gutenberg. The corpus emphasizes long-context settings to assess how a draft model can anticipate the outputs of a large model. Besides the content of the book, the instruction prompt comprises three tasks: (1) produce a concise synopsis of the entire text; (2) select any character or episode and provide a focused analysis of technique and meaning; and (3) reflect on the implications of the novel for contemporary life. The specific books are as follows:

\begin{itemize}
  \item \textit{A Doll’s House} (Gutenberg ID: 2542)
  \item \textit{A Room with a View} (Gutenberg ID: 2641)
  \item \textit{Alice’s Adventures in Wonderland} (Gutenberg ID: 11)
  \item \textit{Frankenstein; Or, The Modern Prometheus} (Gutenberg ID: 84)
  \item \textit{Simple Sabotage Field Manual} (Gutenberg ID: 26184)
  \item \textit{The Great Gatsby} (Gutenberg ID: 64317)
  \item \textit{The Importance of Being Earnest} (Gutenberg ID: 844)
  \item \textit{The Picture of Dorian Gray} (Gutenberg ID: 174)
  \item \textit{The Strange Case of Dr. Jekyll and Mr. Hyde} (Gutenberg ID: 42)
  \item \textit{The Tragedy of Romeo and Juliet} (Gutenberg ID: 1112)
\end{itemize}

\end{document}